\documentclass[10pt,twocolumn,letterpaper]{article}

\usepackage[pagenumbers]{cvpr} 

\usepackage[dvipsnames]{xcolor}
\usepackage{algorithm}
\usepackage{algorithmic}
\usepackage{multirow}

\newcommand{\MyModel}{G-FARS}
\definecolor{cvprblue}{rgb}{0.21,0.49,0.74}
\usepackage[pagebackref,breaklinks,colorlinks,citecolor=cvprblue]{hyperref}
\usepackage[accsupp]{axessibility} 

\def\paperID{} 
\def\confName{CVPR}
\def\confYear{2024}

\title{\MyModel: Gradient-Field-based Auto-Regressive Sampling for 3D Part Grouping}

\author{Junfeng Cheng\\
Imperial College London\\
London, UK\\
{\tt\small junfeng.cheng20@imperial.ac.uk}
\and
Tania Stathaki\\
Imperial College London\\
London, UK\\
{\tt\small t.stathaki@imperial.ac.uk}
}

\begin{document}

\maketitle

\begin{abstract}
This paper proposes a novel task named "3D part grouping". Suppose there is a mixed set containing scattered parts from various shapes. This task requires algorithms to find out every possible combination among all the parts. To address this challenge, we propose the so called Gradient Field-based Auto-Regressive Sampling framework (G-FARS) tailored specifically for the 3D part grouping task. In our framework, we design a gradient-field-based selection graph neural network (GNN) to learn the gradients of a log conditional probability density in terms of part selection, where the condition is the given mixed part set. This innovative approach, implemented through the gradient-field-based selection GNN, effectively captures complex relationships among all the parts in the input. Upon completion of the training process, our framework becomes capable of autonomously grouping 3D parts by iteratively selecting them from the mixed part set, leveraging the knowledge acquired by the trained gradient-field-based selection GNN. Our code is available at: \url{https://github.com/J-F-Cheng/G-FARS-3DPartGrouping}.

\end{abstract}

\section{Introduction}
\label{sec:intro}

\begin{figure*}[t]
    \centering
    \includegraphics[width=1.0\textwidth,trim=55 80 55 105,clip]{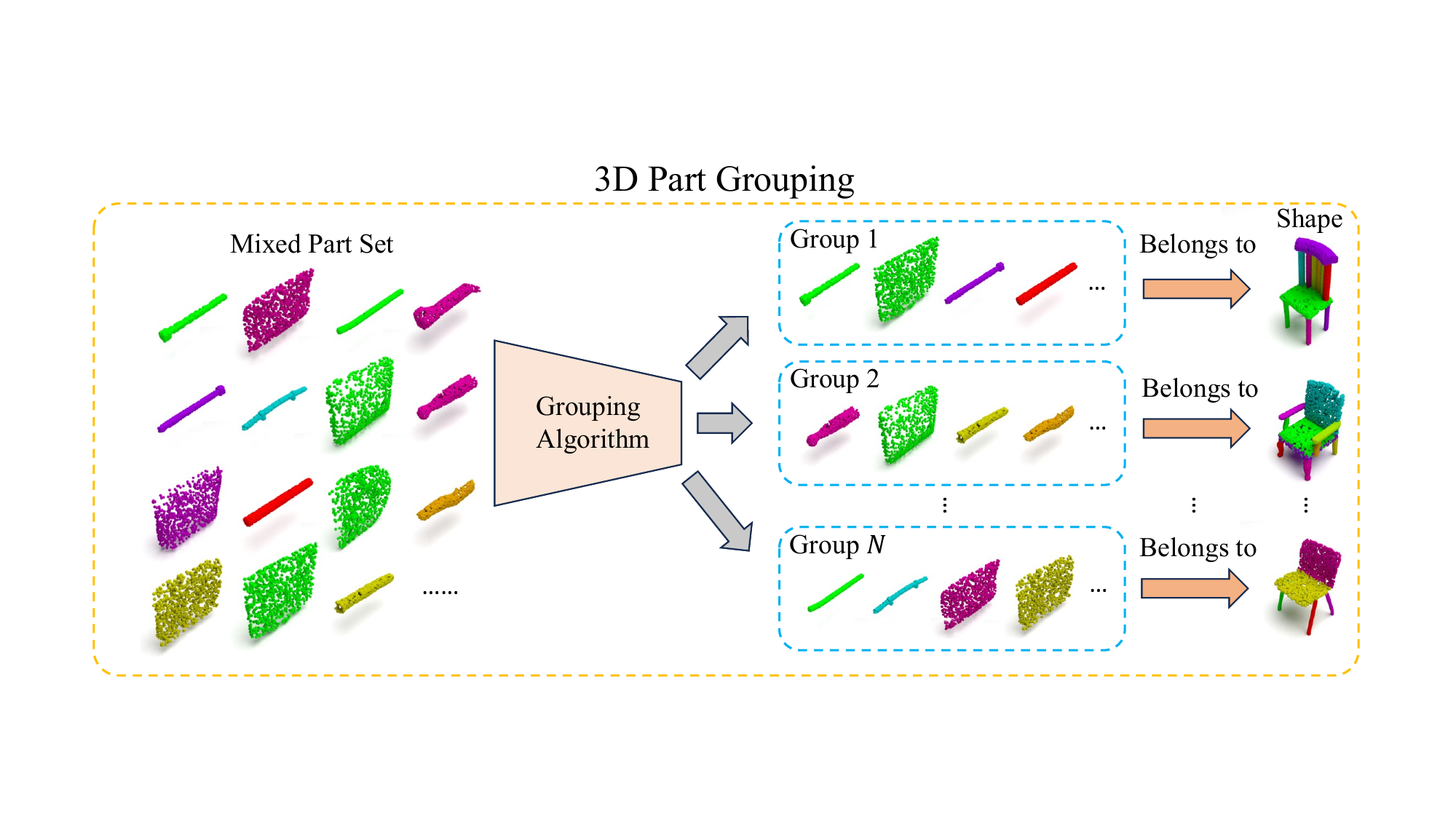}
    \caption{The definition of our proposed 3D part grouping task. Assuming we have a set which contains mixed parts from $N$ shapes. Our goal in this task is to use a grouping algorithm to separate these mixed parts and group them by their respective shapes. }
    \label{fig:3d_grouping_intro}
    
\end{figure*}

Assuming that you purchase multiple unassembled IKEA chairs and carelessly mix all the parts together, it can quickly become a nightmare to sort through and assemble each chair. The task can be especially daunting if the pieces from different chairs are mixed together, making it challenging to identify the correct components for each chair. Similarly, in the field of archaeology, recovering broken cultural relics can be incredibly difficult, as the fragments are often intermingled with the pieces from other relics. In such cases, archaeologists must carefully separate the mixed fragments and piece them together to reconstruct the original relics.
In a similar vein, the field of LEGO automatic assembly requires AI agents to select different combinations of parts from massive LEGO blocks and assemble them into a shape. All of these examples contain two goals: The first goal is to identify the correct combinations from the mixed part set (\emph{i.e.} part grouping) and the second one is to assemble them into reasonable shapes (\emph{i.e.} part assembly). To achieve these two objectives, algorithms must first be capable of comprehending the geometric relationships among all the parts. Next, they should be able to separate the parts by their shapes, and finally, assemble the chosen parts into reasonable shapes.

For the part assembly, previous works have researched some methods for assembling a given group of parts. DGL-Net \cite{zhan2020generative} is the first work to explore the assembly problem without prior instruction. The DGL-Net algorithm can predict the 6-DoF poses for each input part, enabling translation and rotation of the parts to their expected positions. RGL-Net \cite{narayan2022rgl} is another part assembly work that utilizes sequential information among all the input parts. By assembling shapes in a specific order (e.g., top-to-bottom), RGL-Net achieves more accurate assembly. IET \cite{zhang20223d} is a recently proposed algorithm that utilizes an instance encoded transformer and self-attention mechanisms \cite{shaw2018self, vaswani2017attention, shi2020masked, yun2019graph, velivckovic2017graph} to enhance the network's assembly ability. 

However, part grouping still remains an unsolved problem. As previously mentioned, the goal of part grouping is to use algorithms to identify all possible combinations in a mixed part set. To address this, we introduce the 3D part grouping task. The definition of this task is presented in Fig.~\ref{fig:3d_grouping_intro}. Suppose we have a set of mixed parts from $N$ different shapes. The 3D part grouping task mandates the algorithms to process all these parts and categorize them into groups based on their originating shapes.

Our proposed task 3D part grouping is challenging for two main reasons. First, the algorithms must understand the relationships among all the parts. Second, the exact number of potential groups, $N$, is unknown. This uncertainty complicates both the problem formulation and the creation of effective algorithms. To tackle these challenges, we introduce Gradient-Field-based Auto-Regressive Sampling (\MyModel) framework in this paper. This framework integrates a gradient-field-based graph neural network for the encoded parts, aiding in understanding the relationships among all the input parts. Moreover, it can auto-regressively sample new groups from the mixed set, enabling the algorithm to identify all the groups, regardless of the number of potential groups $N$. More details about \MyModel~are in Sec.~\ref{sec:method}.

Our proposed task and algorithm hold significant potential for the industrial world. In manufacturing environments, where parts from multiple products are intermixed, a robotic system utilizing this method can aid in automated sorting, leading to more efficient production lines. The 3D part grouping algorithm can also generalize to other domains. For instance, in recycling facilities, this method could help segregate mixed materials into appropriate categories for processing, thus improving waste management practices.
Our contributions are concluded as follows:

\begin{itemize}
\item We introduce a novel setting termed 3D part grouping, which necessitates algorithms to identify all possible combinations of the input mixed parts.
\item We present the Gradient-Field-based Auto-Regressive Sampling (\MyModel) for the 3D part grouping task. Our framework addresses two primary challenges in this task: 1. understanding the relationships among parts and 2. managing the uncertainty of the potential groups.
\item Utilizing PartNet \cite{mo2019partnet}, we establish benchmarks for our introduced task and showcase the effectiveness of \MyModel~in the 3D part grouping domain.
\end{itemize}

\section{Related Works}

\label{sec:related}

\subsection{Combinatorial Optimization}
Combinatorial Optimization is an area that studies the task of finding the best object from a finite set of objects. The 3D part grouping task aims to identify every possible combination in a given set of parts. From this perspective, the 3D part grouping task aligns with combinatorial optimization problems, and we can glean insights from combinatorial optimization algorithms.

\paragraph{Classic algorithms} Research in combinatorial optimization can be traced back to the 1960s. Some hallmark algorithms include Dijkstra's shortest path algorithm \cite{johnson1973note, chen2003dijkstra, jasika2012dijkstra, jiang2014extending, samah2015modification} and Kruskal's minimum spanning tree algorithm \cite{haripriya2023performance}. These algorithms provided a robust foundation for understanding the structure and intricacies of combinatorial problems.

\paragraph{Heuristic methods} have introduced a new dimension to combinatorial optimization. Genetic algorithms \cite{han2022improved, gen2023genetic, holland1992genetic, forrest1996genetic, srinivas1994genetic}, inspired by natural selection processes, stand as prominent methods for these problems. They employ mechanisms like mutation, crossover, and selection to pinpoint solutions. Another significant method is Ant Colony Optimization \cite{dorigo2006ant, blum2005ant, kavitha2023ant, zhou2022parameter, karimi2023semiaco}, which emulates the behavior of ants when finding a path from their colony to food sources.

\paragraph{Reinforcement learning-based methods} Recently, some reinforcement learning-based methods also demonstrate their ability in combinatorial optimization problems \cite{mazyavkina2021reinforcement, bello2016neural}. The basic idea behind these methods is to learn a policy to search the solution space. 

\paragraph{Graph neural networks} are also making their mark in addressing combinatorial optimization problems. They transpose the problem graph into continuous space, aiding in the prediction of optimal combinatorial solutions. The essence of graph neural networks revolves around the use of message passing and aggregation operations on nodes and edges to unearth and understand the graph's layout. Graph Convolutional Networks (GCNs) are a notable type of graph neural network that applies convolutional operations on graphs, capturing local and global structural nuances \cite{kipf2016semi}. Edge Convolution, another variant, focuses on the features of edges between nodes, amplifying the model's capacity to depict complex graphs \cite{wang2019dynamic}.

\subsection{3D Part Assembly}
Although this work does not consider the task of part assembly, it is still worthwhile to review these works, as they provide insights into identifying the relationships among the input parts.

3D part assembly is a task proposed by Huang et al. \cite{zhan2020generative} which aims to assemble separate parts into a complete shape without any external guidance. The goal of 3D part assembly is to predict a translation vector and a rotation vector for each part \cite{zhan2020generative, narayan2022rgl, zhang20223d, cheng2023score}. We introduce two categories of 3D part assembly algorithms, graph neural network based algorithms and transformer based algorithm:
\paragraph{Graph neural network based methods}
Huang et al. \cite{zhan2020generative} propose Dynamic Graph Learning algorithm to achieve the goal of 3D part assembly task. DGL-Net includes a PointNet \cite{qi2017pointnet} for part feature extraction. They also propose an iterative graph neural network backbone for message passing among all the part features. Besides, they propose dynamic relation reasoning modules to learn the relationship among all the encoded parts. They also have dynamic part aggregation modules for more direct information exchanges among geometrically-equivalent parts. 

RGL-Net, described in \cite{narayan2022rgl}, is also a GNN-based algorithm designed for the part assembly task. The key concept behind RGL-Net is sequential assembly, where the separate parts are assembled in a specific order, such as a top-to-down approach. In particular, the authors employ GRUs \cite{chung2014empirical} to learn the order information, which significantly improves the assembling performance. However, one potential limitation of RGL-Net is that its performance may be suboptimal in non-ordered assembly settings. In other words, when the parts are assembled in a random or unstructured order, RGL-Net may not perform as well as it does in the ordered assembly setting.

\paragraph{Transformer based method}

Zhang et al. \cite{zhang20223d} propose a method that employs a transformer-based framework and self-attention mechanism to model the relationships between different parts, while resolving the ambiguity issue using instance encoding. The method includes four modules: a shared PointNet for feature extraction, a transformer encoder for reasoning the relationships between parts, an MLP predictor for pose estimation, and an instance encoder for handling the ambiguity between parts. The experimental results show that IET achieves better performance than DGL-Net and RGL-Net.

\section{Backgrounds of Score-based Modeling Through Stochastic Differential Equations}
\label{sec:back_sde}

Our proposed \MyModel~is built upon the score-based modeling through stochastic differential equations (SDEs). It is necessary to discuss the basic principles of the score-based modeling before introducing \MyModel.

Score-based modeling through SDEs \cite{song2020score} is a recently proposed technique for generation tasks. The basic idea behind this modeling method is estimating the gradients of the data distribution. The new data can be generated by sampling the estimated data distribution. Score-based models with SDEs are trained to estimate a time-dependent score $S_\theta(\mathbf{x}(t), t)$ of a given probability density function $p_t(\mathbf{x})$, which can be described by $S_\theta(\mathbf{x}(t), t)=\nabla_\mathbf{x}\log p_t(\mathbf{x})$. To successfully achieve a score-based model, we need to set up a diffusion process which can be represented by a continuous-time stochastic process $\{\mathbf{x}(t) \in \mathbb{R}^d \}_{t=0}^T$, where $t \in [0,T]$. We choose a diffusion process such that $\mathbf{x}(0)\sim p_0$ and $\mathbf{x}(T)\sim p_T$, where $p_0$ is the data distribution, $p_T$ is the prior distribution, and both are uncorrelated after the perturbation by the diffusion process. We can describe our diffusion process by using the following equation:
\begin{equation}
d \mathbf{x} = \mathbf{f}(\mathbf{x}, t) d t + g(t) d \mathbf{w}.
\end{equation}
In the above equation, $\mathbf{f}(\cdot, t): \mathbb{R}^d \to \mathbb{R}^d$ is the drift coefficient of the SDE, $g(t) \in \mathbb{R}$ represents the diffusion coefficient, and $\mathbf{w}$ means the standard Brownian motion. To generate new data, we can sample the score function by reversing the diffusion process:
\begin{equation}
  d\mathbf{x} = [\mathbf{f}(\mathbf{x}, t) - g^2(t)\nabla_{\mathbf{x}}\log p_t(\mathbf{x})] dt + g(t) d\bar{\mathbf{w}},
\end{equation}
where $\bar{\mathbf{w}}$ denotes a Brownian motion with time flowing in the reverse direction, and $dt$ is an infinitesimal time step with a negative sign. To successfully train a score-based model, we optimize the following objective function:
\begin{multline}
\label{equ:score_obj}
\min_\theta \mathbb{E}_{t\sim \mathcal{U}(0, T)} [\lambda(t) \mathbb{E}_{\mathbf{x}(0) \sim p_0(\mathbf{x})}\mathbb{E}_{\mathbf{x}(t) \sim p_{0t}(\mathbf{x}(t) \mid \mathbf{x}(0))}\\
[ \|S_\theta(\mathbf{x}(t), t) - \nabla_{\mathbf{x}(t)}\log p_{0t}(\mathbf{x}(t) \mid \mathbf{x}(0))\|_2^2]],
\end{multline}

\section{Method}

\label{sec:method}

This section introduces our proposed G-FARS framework. We start by discussing the working principle of the framework. We then explain how we establish its key component, referred to as the gradient-field-based selection GNN. Following that, we explore the training algorithm for our framework. Finally, we discuss the sampling algorithm within the G-FARS framework.

\begin{figure*}[t]
    \centering
    \includegraphics[width=1.\textwidth,trim=50 60 10 120,clip]{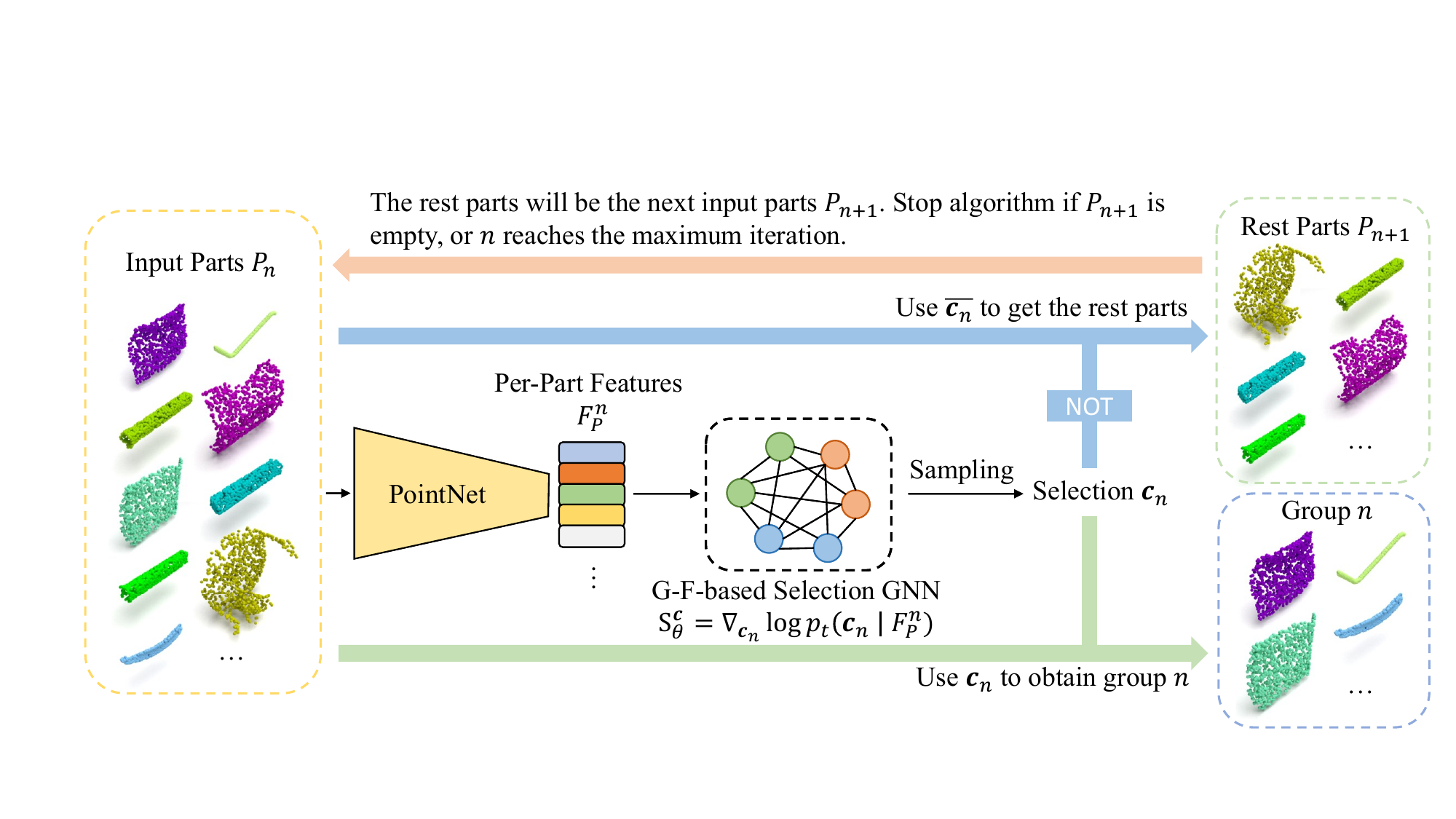}
    \caption{The auto-regressive sampling procedure of our proposed framework. First, we obtain the per-part feature by using a PointNet \cite{qi2017pointnet} to encode all the input parts $P_n$ at the iteration $n$. Then, we use the gradient-field-based (G-F-based) selection GNN to sample a selection vector $\mathbf{c}_n$ to obtain part group $n$, and use $\overline{\mathbf{c}_n}$ to get the rest parts $P_{n+1}$. $P_{n+1}$ will be the next input parts at the next iteration. The auto-regressive sampling stops when $P_{n+1}$ is empty, or $n$ reaches the maximum iteration. }
    \label{fig:ar_sampling}
\end{figure*}

\subsection{\MyModel: Gradient-Field-based Auto-Regressive Sampling Framework}

We propose a model, denoted as \MyModel, to solve the 3D part assembly problem, the workings of which are illustrated in Fig.~\ref{fig:ar_sampling}. Before we discuss our framework, we introduce the definitions of the mathematical symbols first. 
 \vspace{-.2cm}
\paragraph{Mathematical symbol definitions} Let $n$ denote the index for iteration, and $P_n$ represents the mixed part set at iteration $n$. Naturally, $P_0$ is the initial input part set before any processing by the algorithm. Each part in the mixed part set is a point cloud which has the dimension of $1000 \times 3$. The Boolean vector $\mathbf{c}_n$ is used for part selection. At iteration $n$, this vector determines how to select parts from the mixed part set $P_n$ to form a new group $n$. The ultimate goal of our algorithm is to identify all possible groups.
\vspace{-.2cm}
\paragraph{Working principle} Our framework consists of two key components: a PointNet and a gradient-field-based selection graph neural network (GNN). The PointNet is employed to encode all the input parts. The gradient-field-based selection GNN is constructed with Edge Convolution layers \cite{wang2019dynamic}, which enables the framework to understand the relationships among all the input parts. We can use this network to sample new selection vectors for the encoded parts. Our framework operates as an auto-regressive algorithm. Assuming we are at iteration $n$, we begin with using the PointNet to obtain the encoded per-part features $F_P^n$. Following that, our gradient-field-based selection GNN is used to sample a selection vector $\mathbf{c}_n$. The sampled vector $\mathbf{c}_n$ allows us to identify group $n$, while the complementary vector $\overline{\mathbf{c}_n}$ (obtained via a bitwise NOT operation) determines the unselected parts. These unselected parts then form the input parts $P_{n+1}$ for the subsequent iteration $n+1$.
\vspace{-.2cm}
\paragraph{Auto-Regressive Sampling} As stated in the Introduction section, one critical issue in the 3D part grouping task is that the number of groups $N$ is not certain. Our framework is able to solve this problem as it can auto-regressively sample new groups from the mixed part set. As stated in the working principle part, our framework is able to sample a new group at each iteration. In this case, our algorithm does not care the number $N$ in the mixed part set, and it only stops when the mixed part set is cleaned or the algorithm reaches the maximum iterations. 

\subsection{Gradient-Field-based Selection Graph Neural Network}
We have already discussed the main working principle of \MyModel. The problem remaining here is how to obtain the key component (\emph{i.e.}, the gradient-field-based selection graph neural network $S_\theta^{\mathbf{c}}$) in our proposed framework. In our designed GNN, the nodes are the per-part features $F_p^n$ encoded by the PointNet. These nodes are fully connected to facilitate message passing.

\paragraph{The learning objective} Designing a suitable learning objective for the gradient-field-based selection graph is key to achieving success in auto-regressive sampling. Recalling the principle of \MyModel, we expect our designed model to predict the correct selection for the input parts $P_n$ at iteration $n$. To achieve this, we design our gradient-field-based GNN to learn the distribution of selections conditioned on the input parts. Mathematically, our gradient-field-based selection GNN learns a conditional probability $p(\mathbf{c}_n \mid F_p^n)$. After the training process, the GNN learns all the ways of selecting for the mixed part set. In this case, at each iteration, we can use the trained GNN to sample a correct selection for the mixed part set. 

\subsection{Training} 
After confirming our learning objective, our next task is to achieve our goal of estimating the distribution $p(\mathbf{c}_n \mid F_p^n)$. Specifically, our gradient-field-based selection GNN is designed to approximate the gradients of the target log conditional probability density. Mathematically, we expect our model to satisfy $S^\mathbf{c}_\theta = \nabla_\mathbf{c} \log p_t(\mathbf{c}_n \mid F_P^n)$.

To achieve this goal, we basically follow the training steps proposed by Song et al. \cite{song2020score}, and we have discussed this in the Backgrounds section (Sec.~\ref{sec:back_sde}). However, we still need to modify the training objective function to the conditional form. We present the general form of the loss function in the following equation:
\begin{multline}
\hspace{-.38cm}
\label{equ:score_obj_con}
\mathcal{L}(\mathbf{x}, \mathbf{y}) = \lambda(t) 
\|S_\theta^\mathbf{c}(\mathbf{x}(t), t) - \nabla_{\mathbf{x}(t)}\log p_{0t}(\mathbf{x}(t) \mid \mathbf{x}(0), \mathbf{y})\|_2^2,
\end{multline}
where $\mathbf{x}(0)$ is the original data, $\mathbf{x}(t)$ is the perturbed data, $\mathbf{y}$ represents the condition and $t$ is time index. The next step is to train \MyModel~ through Algorithm~\ref{algorithm:training}. In the training algorithm, we first use PointNet to get the per-part feature $F_P^n$. Following that, we sample time index $t$ from the uniform distribution $\mathcal{U}(0, T)$. Then the $\mathbf{c}_n(t)$ can be obtained through the perturbation. After that, we calculate the loss function and perform back propagation. At the end of the iteration, we optimize the parameters for both $PointNet$ and $S^\mathbf{c}_\theta$. 
\begin{algorithm} 
\caption{The training algorithm}
\label{algorithm:training}
\textbf{Input: } Training dataset $\mathcal{D}_{train}$ \\
\textbf{Parameter: } $T$, $N$ epochs \\
\textbf{Output: } $S^\mathbf{c}_\theta$, $PointNet$
\begin{algorithmic}[1]
\FOR{$N$ epochs}
    \FOR{each $P_n, \mathbf{c}_n$ in $\mathcal{D}_{train}$}
        \STATE $F_P^n \leftarrow PointNet(P_n)$;
        \STATE Sample $t\sim \mathcal{U}(0, T)$; 
        \STATE Add perturbation on $\mathbf{c}_n(0)$ to obtain $\mathbf{c}_n(t)$;
        \STATE Calculate loss $\mathcal{L}(\mathbf{c}_n, F_P^n)$;
        \STATE Perform back propagation;
        \STATE Optimize the parameters of $PointNet$ and $S^\mathbf{c}_\theta$;
    \ENDFOR
\ENDFOR
\RETURN G-F Selection GNN $S^\mathbf{c}_\theta$, $PointNet$;

\end{algorithmic}
\end{algorithm}

\subsection{Sampling Algorithm}
\begin{algorithm}
\caption{Predictor-Corrector sampler}
\label{algorithm:PC_sampler}
\textbf{Input}: $F_P^n$ \\
\textbf{Parameter}: $T$, $\sigma$, $N$, $C$ \\
\textbf{Output}: the sampled result $\mathbf{c}_n(0)$
\begin{algorithmic}[1]
\STATE Sample $\mathbf{c}_n(T) \sim \mathcal{N}(\mathbf{0}, \frac{1}{2 \log \sigma}(\sigma^{2T} - 1)\mathbf{I})$.

\FOR{$n \leftarrow N-1$ to $0$}{
    
	\STATE $t_p \leftarrow \frac{(n+1)T}{N}$ $t \leftarrow \frac{nT}{N}$
	
	\FOR{$i \leftarrow 1$ to $C$}{
	
	\STATE $\mathbf{c}_n(t_p) \leftarrow Corrector(\mathbf{c}_n(t_p),F_P^n)$
	}
         \ENDFOR
	\STATE $\mathbf{c}_n(t) \leftarrow Predictor(\mathbf{c}_n(t_p), F_P^n) $

}
 \ENDFOR
 \RETURN the sampled result $\mathbf{c}_n(0)$

\end{algorithmic}
\end{algorithm}

As stated above, our framework auto-regressively samples new selections to choose parts from the given part set. In this case, a suitable sampler for our framework is the key to achieve high performance of 3D part grouping. We use Predictor-Corrector (PC) \cite{song2020score} as our sampler. We demonstrate a simplified version in Algorithm~\ref{algorithm:PC_sampler}. This method ensures that the generated samples are close to the desired distribution. Since our application requires a conditional PC sampler, we have provided the algorithm in~\ref{algorithm:PC_sampler}. 
After the sampling procedure, we use a threshold $T_h=0.5$ to transform the selection vector $\mathbf{c}_n(0)$ to a boolean vector.

\section{Experiment}
\subsection{Datasets} 

\begin{figure}[h]
    \centering
    \includegraphics[width=.4\textwidth,trim=220 0 400 20,clip]{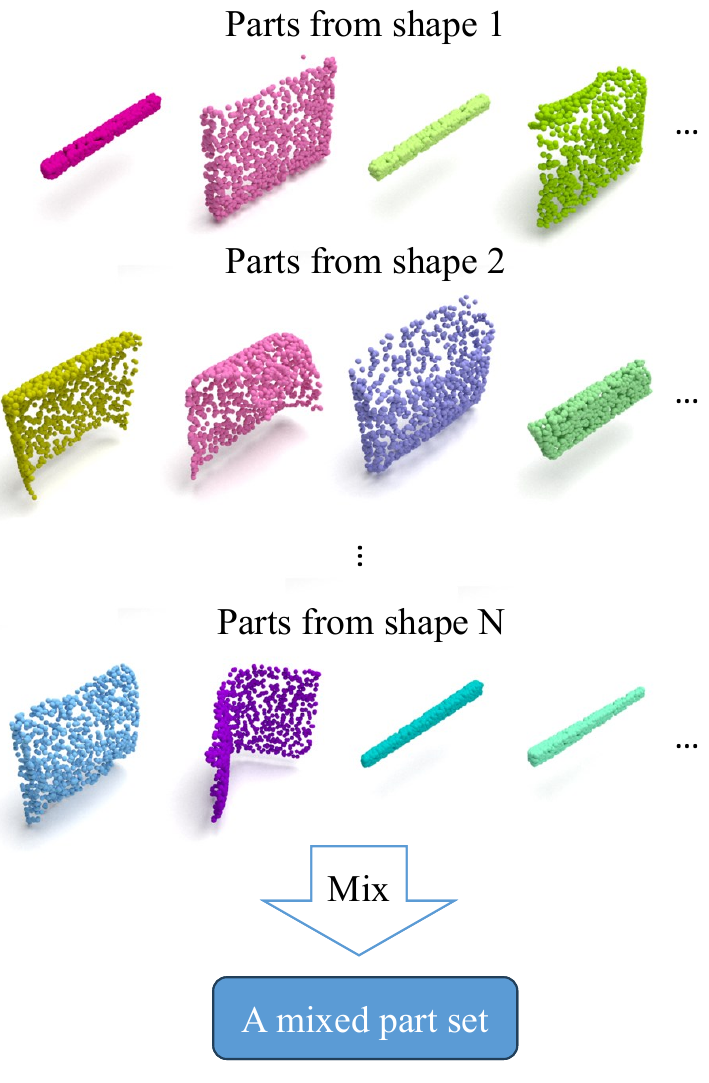}
    \caption{The random mixing method to create a mixed part set. We randomly selected $N$ shapes from the PartNet dataset. We then mix all the parts into a single set. The sequence of parts are shuffled.}
    \label{fig:create_dataset}
\end{figure}

As 3D processing task, PartNet \cite{mo2019partnet} can be a good option for us to conduct our experiments. We apply 6,323 chairs, 8,218 tables and 2,207 lamps from PartNet \cite{mo2019partnet} in our experiments. We apply random mixing method to create our mixed part sets. We illustrate our mixing method in Fig.~\ref{fig:create_dataset}. First, we randomly selected $N$ shapes from the PartNet dataset ($N$ is also a random number). Next, we mixed all the parts into a single part set. Finally, we shuffle all the parts to obtain our mixed set. For more statistics of our mixed datasets, please refer to our Supplementary.
\label{sec:datasets}

\begin{figure*}[t]
    \centering
    \includegraphics[width=1.\textwidth,trim=50 20 50 60,clip]{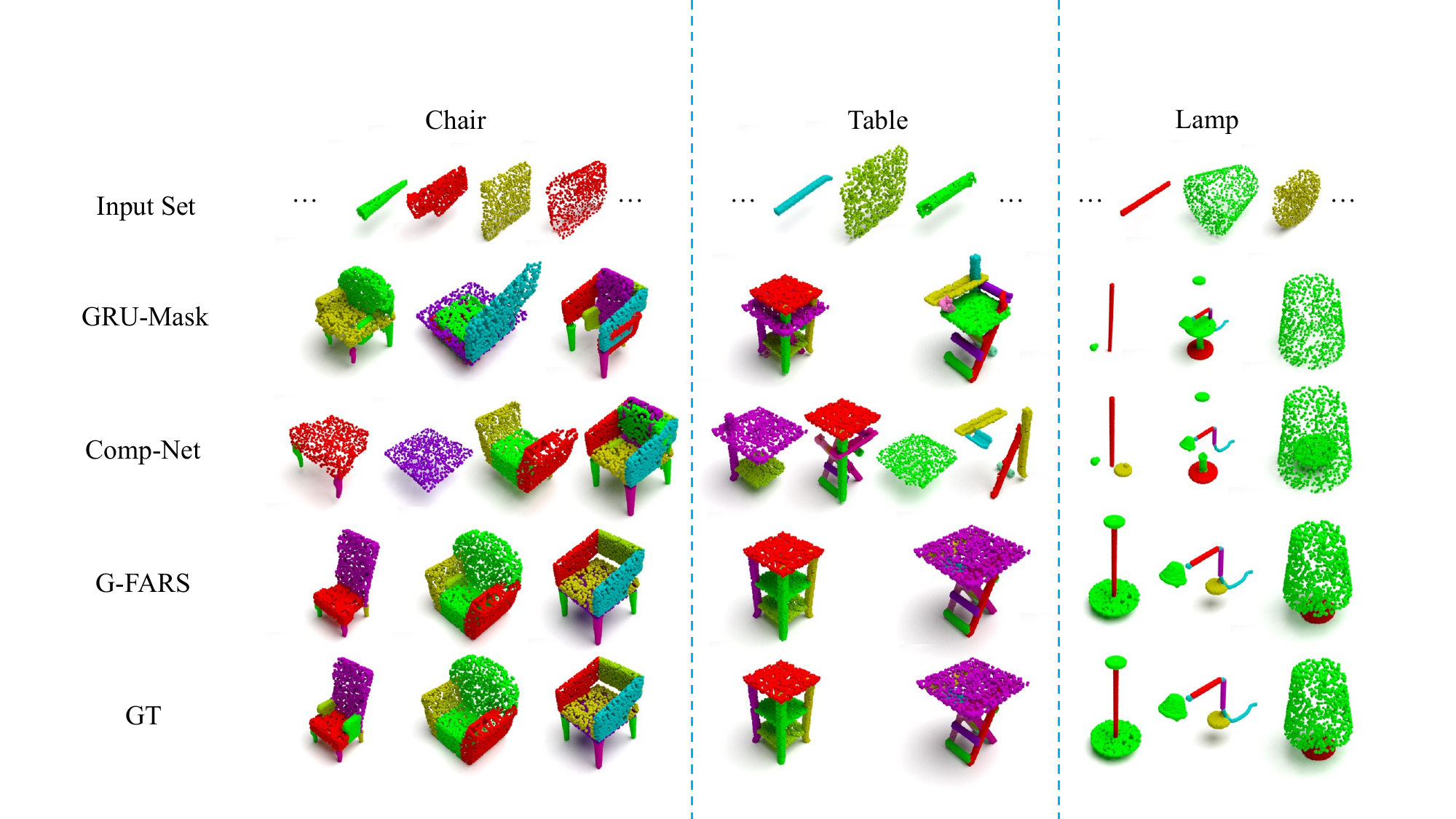}
    
    \caption{The qualitative comparisons. To intuitively demonstrate the effect of grouping, we rotate and translate all the parts by using their ground truth poses after the grouping procedure. The results show that only \MyModel~is able to correctly group the 3D parts. Some baselines even predict an incorrect number of groups (\emph{e.g.}, Comp-Net). Due to the page limit, we here only present the results of \MyModel~and the two most competitive baselines (\emph{i.e.}, GRU-Mask and Comp-Net). We have included the full comparison and an additional qualitative comparison figure in the supplementary materials. }
    \label{fig:qualitative}
    
\end{figure*}

\subsection{Evaluation Metrics} 
We use precision, recall and F1 score to evaluate the performance of the algorithms in 3D part grouping task. 

\paragraph{Definitions for TP, FP and FN in 3D part grouping task} Before the evaluation, it is necessary to define the TP, FP and FN in 3D part grouping task. We here assume we have one predicted group and the corresponding ground truth group. TP, FP and FN are defined as:
\begin{itemize}
    \item TP: The parts selected by the predicted group are also in the ground truth group.
    \item FP: The parts selected by the predicted group are NOT in the ground truth group.
    \item FN: The parts which are NOT selected by the predicted group are in the ground truth group.
\end{itemize}

\paragraph{Single set average VS overall average} To fully investigate the performance of the algorithms, we provide two averaging methods in the evaluation. We present the definitions as follows:

\begin{itemize}
    \item Single set average: We calculate the TP, FP and FN in one part set, and calculate and record the corresponding precision, recall and F1 score of this single part set. After all the input part sets are processed, we calculate the average precision, recall and F1 score.
    \item Overall average: We calculate the TP, FP and FN for all the part sets and record them. After all the input part sets are processed, we calculated precision, recall and F1 score based on the accumulated TP, FP and FN. 
\end{itemize}

\paragraph{Jaccard similarity matching} In the evaluation procedure, a critical issue is that the sequence of the predicted part groups is not certain. In this situation, for fair evaluation, we apply Jaccard similarity to match the ground truth group with the predicted group. The formula of Jaccard similarity is shown as follows:
\begin{equation}
    J(P, G) = \frac{|P \cap G|}{|P \cup G|}
\end{equation}
where $P$ is the predicted group, and $G$ is the ground truth group. In our task, $|P \cap G|$ represents the number of parts which are included by both $P$ and $G$, while $|P \cup G|$ represents the total number of unique parts in both $P$ and $G$. In our evaluation program, each ground truth group is matched with the predicted group which has the highest Jaccard similarity.

\subsection{Baselines}

As 3D part grouping is a newly proposed task, there are no existing baselines specific to this task. Therefore, we propose some alternative methods for comparisons:
\begin{itemize}
    \item \textbf{GRU-Mask:} We propose GRU-Mask which uses a GRU to generate a binary mask for the given part set. This mask indicates how to select parts to form groups. GRU-Mask is inspired by RNN-based methods \cite{zou20173d, wu2020pq}. More details about GRU-Mask can be found in the supplementary.
    \item \textbf{Comp-Net:} Comp-Net is another method proposed by us, capable of classifying whether two parts can be grouped to form a shape. Our idea is inspired by RL-based approaches \cite{tulsiani2017learning, luo2020learning}. We provide more information about Comp-Net in our supplementary.
    \item \textbf{Variants of G-FARS:} Inspired by \cite{shi2021learning, luo2021score, baranchuk2021label, chen2023mixed, mo2023dit}, we propose three variants of G-FARS: G-FARS-CG, G-FARS-R, and G-FARS-T. We use the same training and inference algorithms (\emph{i.e.,} Auto-Regressive sampling) as \MyModel~in these variants. 
    G-FARS-CG applies a GNN for message passing among all the encoded parts. Besides, it also includes a separate MLP to learn the score function for part selection. G-FARS-R and G-FARS-T use a ResNet \cite{he2016deep} and a transformer \cite{vaswani2017attention} to learn the selection score function respectively. 
    For these variants, we employ a score function modeling way that differs from the one used in \MyModel, which is introduced in the supplementary materials.

\end{itemize}

\subsection{Experimental Details}
\paragraph{Hardware and software details} We conduct all the experiments on a personal computer with CPU R9 3900x and RTX3090. The memory of the PC is 32 GB. The operating system is Ubuntu Linux. 
\paragraph{About hyper-parameters} Our framework contains many hyper-parameters. For the sampler and the sampling steps, we discuss them in Table \ref{tab:ablation}. For other details about hyper-parameters, please refer to our supplementary.

\subsection{Comparisons and Discussions}

We present the quantitative comparisons in Table~\ref{tab:quantitative}. Overall, our framework \MyModel~outperforms other baselines by a large margin on all datasets. This proves the effectiveness of our proposed \MyModel. We also demonstrate the qualitative results in Fig. \ref{fig:qualitative} (the full comparison is in the supplementary). To intuitively show our performance of the grouping results, we rotate and translate the parts in all the groups with their ground truth poses after the grouping procedure. \textbf{Please Note that the rotation and translation are only used for demonstration purpose. We DO NOT use the ground truth poses information in the grouping procedure or the training procedure.} The figure shows that our \MyModel~is able to group most of the mixed part sets. However, the other baselines hardly manage to predict the correct groups accurately. 
Besides these experiments, we also demonstrate category mixing testing, and generalization testing to unseen categories in the supplementary. These tests further prove the effectiveness of \MyModel.

\begin{table*}[t]
    \centering
    \begin{tabular}{c|c|cc|cccc}
    \toprule[1.5pt]
         Metrics & Category & GRU-Mask & Comp-Net & G-FARS-CG & G-FARS-R & G-FARS-T & G-FARS \\
         \midrule

\multirow{3}{*}{Precision ↑}& Chair & 0.641 / 0.608 & 0.696 / 0.629 & 0.592 / 0.532 & 0.586 / 0.522 & 0.6 / 0.539 & \textbf{0.828} / \textbf{0.793} \\
& Table & 0.647 / 0.601 & 0.758 / 0.694 & 0.62 / 0.555 & 0.608 / 0.545 & 0.596 / 0.528 & \textbf{0.848} / \textbf{0.811} \\
& Lamp & 0.613 / 0.576 & 0.732 / 0.661 & 0.691 / 0.616 & 0.684 / 0.619 & 0.718 / 0.637 & \textbf{0.751} / \textbf{0.711} \\
\midrule
\multirow{3}{*}{Recall ↑}& Chair & 0.67 / 0.652 & 0.687 / 0.674 & 0.449 / 0.441 & 0.476 / 0.473 & 0.452 / 0.446 & \textbf{0.753} / \textbf{0.744} \\
& Table & 0.689 / 0.667 & 0.694 / 0.677 & 0.448 / 0.439 & 0.463 / 0.453 & 0.492 / 0.484 & \textbf{0.798} / \textbf{0.784} \\
& Lamp & 0.657 / 0.633 & 0.641 / 0.634 & 0.547 / 0.531 & 0.568 / 0.543 & 0.509 / 0.491 & \textbf{0.728} / \textbf{0.71} \\
\midrule
\multirow{3}{*}{F1 Score ↑}& Chair & 0.651 / 0.629 & 0.678 / 0.651 & 0.499 / 0.483 & 0.512 / 0.496 & 0.503 / 0.488 & \textbf{0.781} / \textbf{0.768} \\
& Table & 0.662 / 0.632 & 0.712 / 0.685 & 0.507 / 0.49 & 0.514 / 0.495 & 0.525 / 0.505 & \textbf{0.814} / \textbf{0.797} \\
& Lamp & 0.626 / 0.603 & 0.666 / 0.647 & 0.592 / 0.571 & 0.603 / 0.578 & 0.574 / 0.555 & \textbf{0.73} / \textbf{0.711} \\
\bottomrule
    \end{tabular}
    \caption{The quantitative comparisons among all the algorithms. We show both the single set average (before the slash) and overall average (after the slash) results in the table. }
    
    \label{tab:quantitative}
\end{table*}

\subsection{Ablation Study}
\label{sec:ablation}
To further prove the effectiveness of our algorithm, we conducted ablation studies. These experiments were carried out on the Chair dataset. Our ablation experiments comprise two parts: architecture ablation (see Table~\ref{tab:struct}) and sampling ablation (see Table~\ref{tab:ablation}).

In the architecture ablation, we removed two key components from our \MyModel. For \MyModel~w/o GF, we use a deterministic loss (\emph{i.e.,} Binary Cross Entropy) instead of a score-matching loss (refer to Equation~\ref{equ:score_obj_con}) \cite{song2020score} to train a network with the same architecture, adding a Sigmoid activation at the output. In the case of \MyModel~w/o Graph, we replace the GNN with an MLP to learn $S_\theta^\mathbf{c}$. The results in Table~\ref{tab:struct} prove the effectiveness of both modules.

For the sampling ablation, we propose a variant of our \MyModel~framework, where the sampler in our framework is replaced by Euler-Maruyama (EM) sampler \cite{song2020score}. The table shows that the performance of PC sampler is better than the performance of the EM sampler. The main reason why PC sampler outperforms the EM sampler is that the PC sampler uses both the numerical SDE solver and the Langevin MCMC as the corrector, while the EM sampler only contains numerical SDE solver. The Langevin MCMC approach can help the algorithm to reduce the error produced by the numerical SDE solver \cite{song2020score}. We also test the best sampling steps for both samplers, and we find that the best sampling steps for PC sampler is 500. 
\begin{table}[h]
    \centering
    \small
    \begin{tabular}{cccc}
    \toprule[1.5pt]
          & Precision & Recall & F1 Score\\
          \midrule
         w/o GF & 0.706 / 0.568 & 0.317 / 0.316 & 0.386 / 0.406 \\
         w/o Graph & 0.578 / 0.52 & 0.479 / 0.472 & 0.512 / 0.495  \\
         G-FARS & \textbf{0.828} / \textbf{0.793} &\textbf{0.753} / \textbf{0.744} & \textbf{0.781} / \textbf{0.768} \\
         \bottomrule
    \end{tabular}
    \caption{The results of architecture ablation. We show both the single set average (before the slash) and overall average (after the slash) results in the table.}
    \label{tab:struct}
\end{table}

\begin{table}[h]
    \centering
    \small
    \begin{tabular}{p{0.8cm}p{0.45cm}ccc}
    \toprule[1.5pt]
  Sampler &Step  & Precision & Recall & F1 Score \\
   \midrule
\multirow{6}{*}{EM}&100 & 0.659 / 0.607 & 0.566 / 0.552 & 0.599 / 0.578 \\
&200 & 0.825 / 0.791 & 0.719 / 0.711 & 0.76 / 0.749 \\
&300 & 0.814 / 0.779 & 0.724 / 0.716 & 0.758 / 0.746 \\
&400 & 0.822 / 0.789 & 0.719 / 0.71 & 0.758 / 0.747 \\
&500 & 0.825 / 0.795 & 0.716 / 0.708 & 0.76 / 0.749 \\
&600 & 0.824 / 0.785 & 0.719 / 0.711 & 0.76 / 0.746 \\
\midrule
\multirow{6}{*}{PC}&100 & 0.658 / 0.608 & 0.568 / 0.559 & 0.6 / 0.582 \\
&200 & 0.827 / 0.792 & 0.736 / 0.729 & 0.77 / 0.759 \\
&300 & 0.816 / 0.782 & 0.734 / 0.724 & 0.764 / 0.752 \\
&400 & 0.826 / 0.791 & 0.74 / 0.733 & 0.771 / 0.761 \\
&500 & 0.828 / 0.793 &\textbf{0.753} / \textbf{0.744} & \textbf{0.781} / \textbf{0.768} \\
&600 & \textbf{0.833} / \textbf{0.797} & 0.747 / 0.74 & 0.779 / 0.767 \\
\bottomrule
    \end{tabular}
    \caption{The ablation study for samplers and sampling steps. The ablation is conducted on the Chair dataset. We show both the single set average (before the slash) and overall average (after the slash) results in the table. }
    \label{tab:ablation}
\end{table}

\subsection{Noisy Part Removal}
\label{sec:downstream}

\begin{figure}[h]
    \centering
    \includegraphics[width=.45\textwidth,trim=100 190 400 105,clip]{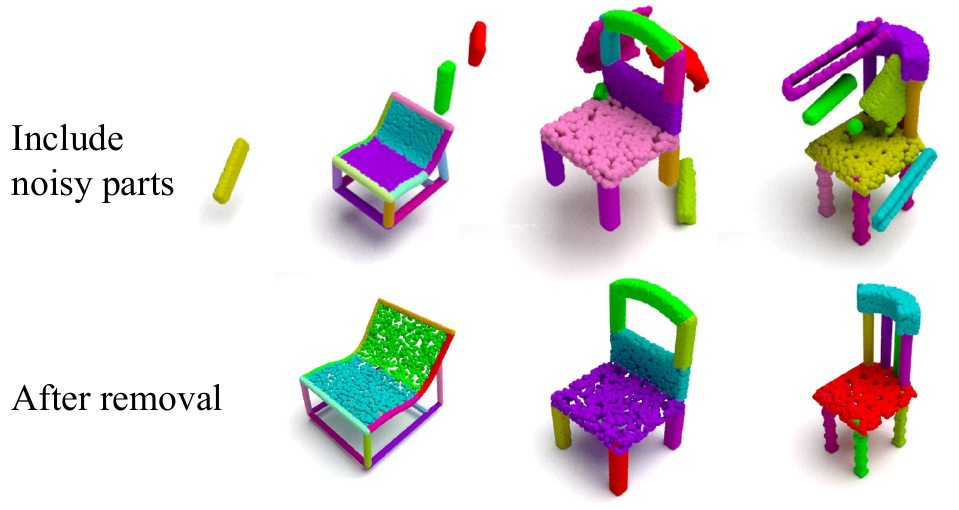}
    \caption{Three examples of noisy part removal. }
    \label{fig:part_removal}
\end{figure}

We surprisingly find that our algorithm is able to achieve the application of noisy part removal in a zero-shot manner. We demonstrate this application in Fig.~\ref{fig:part_removal}. Assume you have a set of parts which belongs to a chair. However, you carelessly mix some noisy parts into this set, and you want to remove these parts. Our framework \MyModel~can achieve your goal. The framework can directly output the correct selection for the parts of your chair. 

\vspace{-.3cm}
\section{Conclusion and Future Work}
In conclusion, we have introduced a novel task termed "3D part grouping", which entails identifying all possible combinations from a mixed set. To address this, we began by randomly mixing shapes from PartNet \cite{mo2019partnet} to construct our training and testing datasets. Subsequently, we unveiled a unique framework named \MyModel~to fulfill our grouping objective. We validated our method using a series of benchmarks, illustrating that our algorithm displays commendable performance on the introduced task.

\paragraph{Future Work} One constraint of our study is its confinement to virtual environments. Moving forward, we aim to investigate the viability of our proposed technique in real-world settings. As an example, our algorithm could be integrated into a robotic system, allowing robots to discern all feasible combinations from a real mixed part set.

\section*{Supplementary}
\appendix

\begin{figure*}[h]
    \centering
    \includegraphics[width=0.85\textwidth,trim=10 20 20 36,clip]{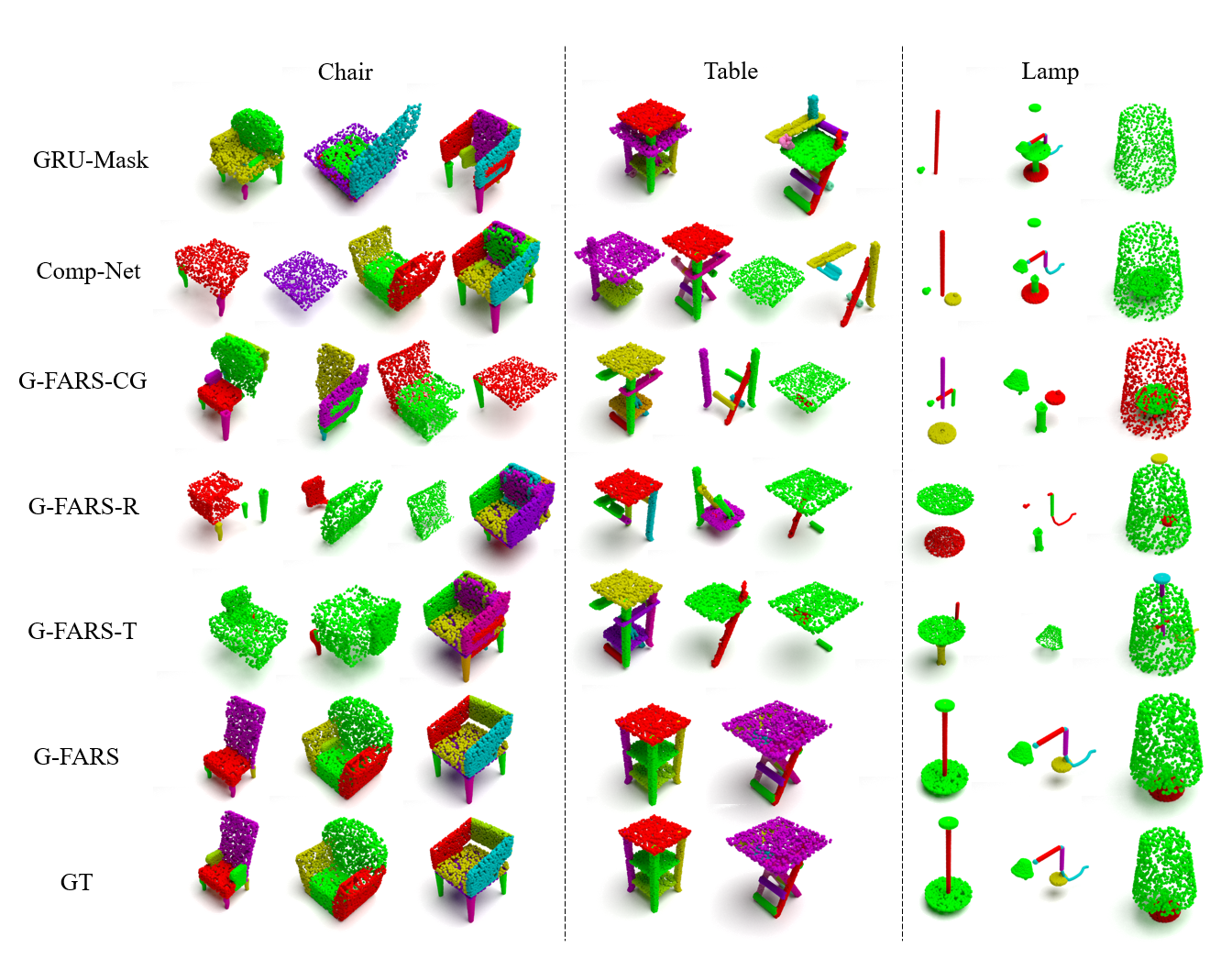}
    \caption{The full comparison for Fig. 4 of the main paper.}
    
    \label{fig:all_quali}
\end{figure*}

\section{Datasets}
As stated in the main paper, we applied three types of shapes, chair, table and lamp from PartNet \cite{mo2019partnet} dataset in our experiments. We use the random mixing method discussed in the main paper to create our training and testing data. In this section, we discuss more details about our datasets.

\paragraph{Data statistics}
In the main paper, we have described our approach to generate 3D part grouping datasets. We provide detailed statistics of our 3D part grouping datasets in the table below:

\begin{table}[h]
    \centering
    \small
    \begin{tabular}{l|ccc}
    \toprule[1.5pt]
        Dataset & Mixed Num. & Train Mixed Set & Test Mixed Set \\
        \midrule
        \multirow{3}{*}{Chair} & 2  & 2381 & 329 \\
        \cmidrule{2-4}
         & 3 & 1032 & 139 \\
         \cmidrule{2-4}
         & total & 3413 & 468 \\
         \midrule
         \multirow{3}{*}{Table} & 2  & 3160 & 471\\
        \cmidrule{2-4}
         & 3 & 1351 & 208 \\
         \cmidrule{2-4}
         & total & 4511 & 679 \\
         \midrule
         \multirow{3}{*}{Lamp} & 2  & 1352 & 209\\
        \cmidrule{2-4}
         & 3 & 620 & 90 \\
         \cmidrule{2-4}
         & total & 1972 & 299 \\
         \bottomrule
         
    \end{tabular}
    \caption{The detailed statistics of our mixed part datasets.}
    \vspace{-.5cm}
    \label{tab:dataset_stat}
\end{table}

\begin{figure*}[h]
    \centering
    \includegraphics[width=.9\textwidth,trim=100 20 100 20,clip]{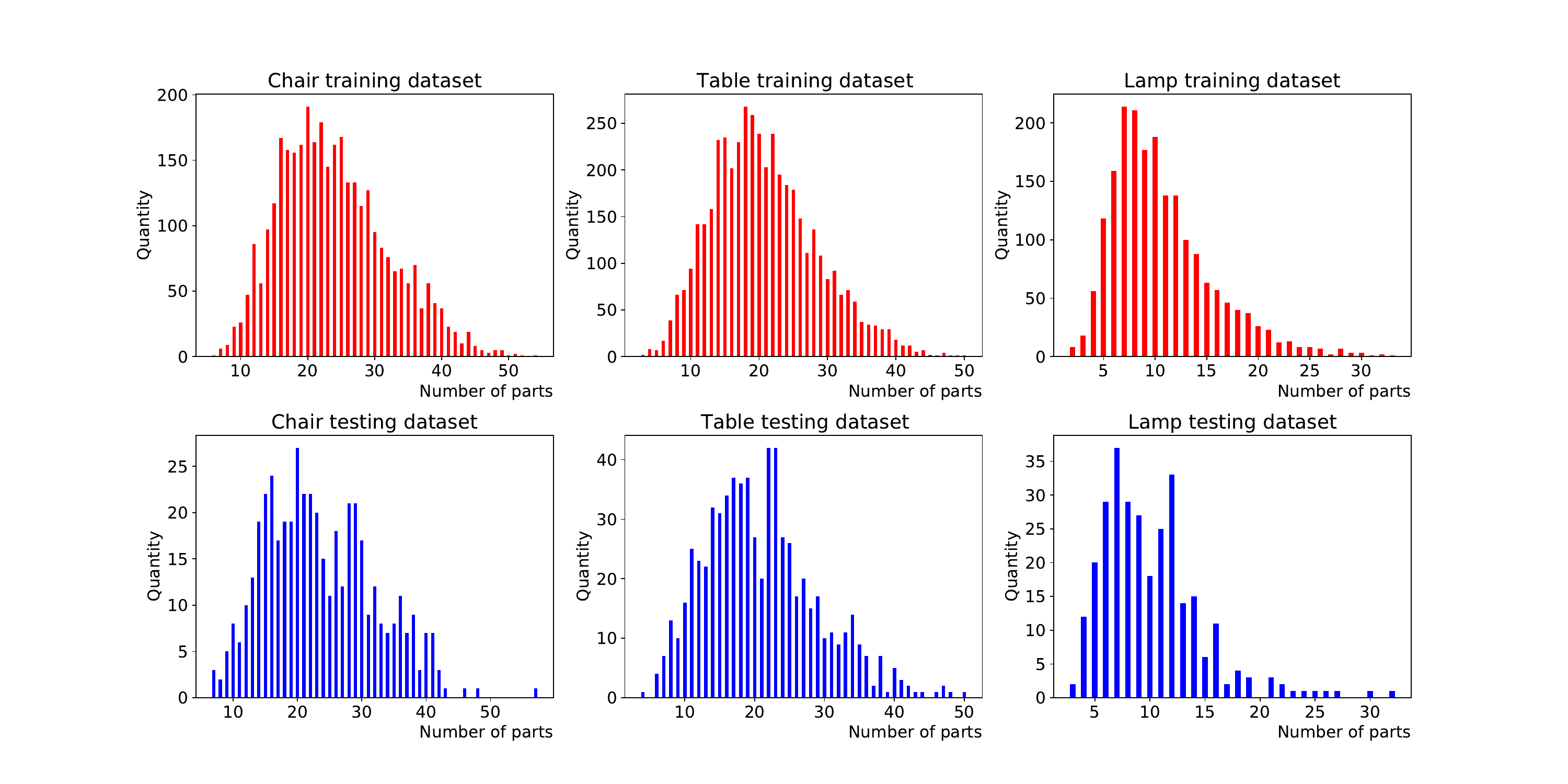}
    \caption{The statistics for the number of parts in the mixed part set. The horizontal axis is the number of parts in one mixed part set, and the vertical axis represents the quantity of the mixed part sets which include the corresponding number of parts. }
    \label{fig:part_stat}
\end{figure*}

\paragraph{Statistics for the Number of Parts} We present the detailed statistics for the number of parts in the constructed mixed part sets in Fig.~\ref{fig:part_stat}. In this figure, the statistics for both training and testing datasets of the three shapes are shown. The horizontal axis represents the number of parts in a single mixed part set, while the vertical axis indicates the quantity of the corresponding part sets in the datasets. For the chair and table datasets, the maximum number of parts exceeds 50; however, for the lamp dataset, this number is only 30. Most part sets in the chair and table datasets contain 10 to 35 parts, while the majority of part sets in the lamp dataset include 5 to 17 parts.

\begin{figure*}[t]
    \centering
    \includegraphics[width=0.85\textwidth,trim=10 10 10 10,clip]{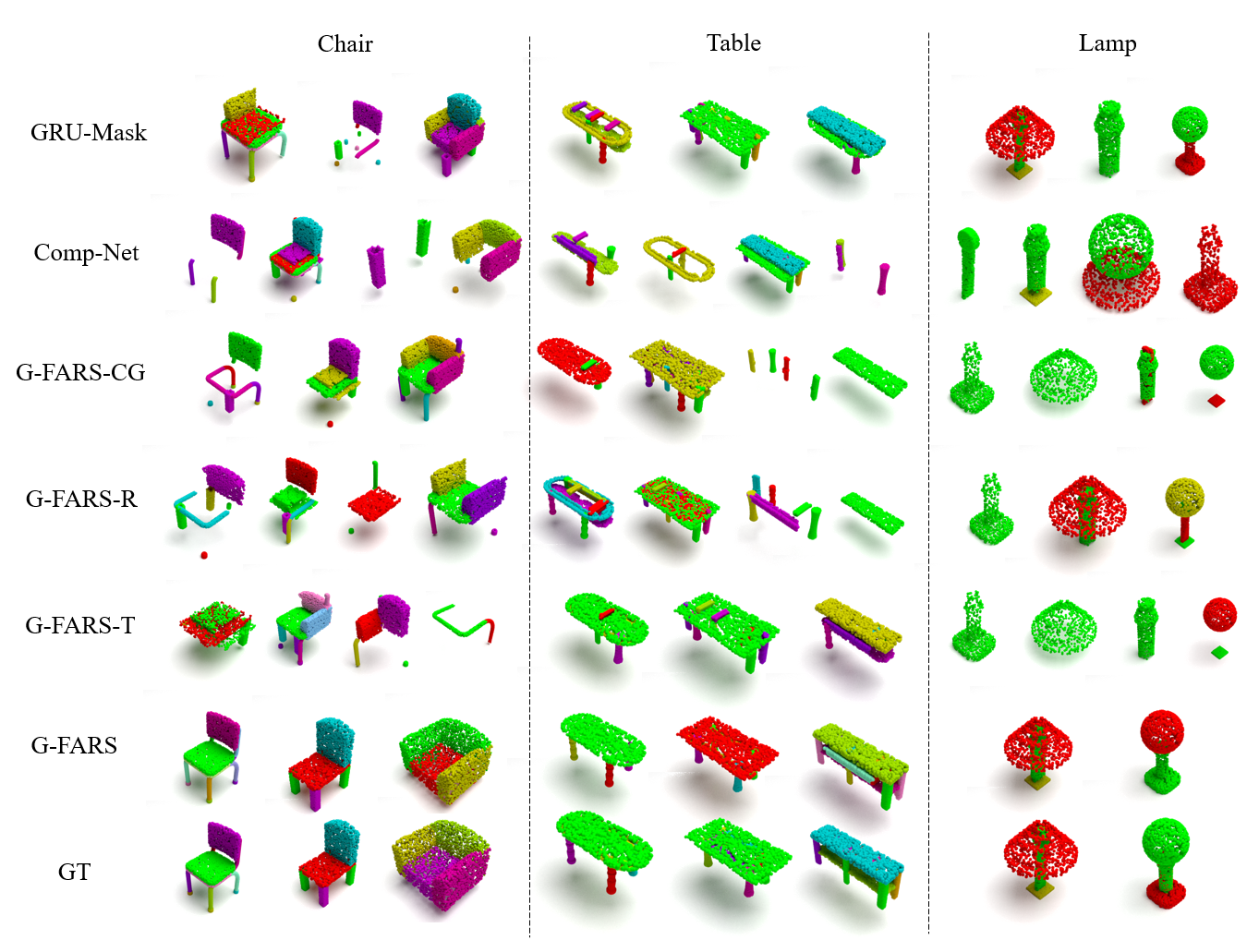}
    \caption{More qualitative comparisons on chair, table and lamp datasets.}
    
    \label{fig:all_quali_2}
\end{figure*}

\section{More Details about Baselines}
We roughly introduce our baselines in the main body, and more details about them are discussed here.
\subsection{GRU-Mask}
In this baseline, we employ the same 3D encoding technique (\emph{i.e.}, PointNet \cite{qi2017pointnet}) as \MyModel~uses. Following the PointNet, we utilize a GRU to sequentially encode the input parts. The GRU enables us to capture the relationships among all parts. Finally, we apply an MLP to generate a mask that represents all the selection methods for the parts.

The quantitative and qualitative comparisons demonstrate that this method achieves the goal of 3D part grouping to some extent. It can generate all the selection vectors without relying on auto-regressive inference. However, a drawback of this method is the necessity to predetermine the output size of the MLP during network design, which limits the number of groups to this predetermined number.

\subsection{Comp-Net}
The idea of Comp-Net is to compare two parts and identify whether they can be grouped together. To implement this concept, we employ a 'dual PointNet' structure. The first PointNet encodes all the input parts, while the second one is tasked with the goal of part comparison. The output of the second PointNet is a boolean value, indicating whether two parts can be grouped together.

Based on the results discussed in the main paper and supplementary materials, we see that this method is a feasible approach for the 3D part grouping task. However, a disadvantage of this approach is that Comp-Net is only trained to compare any two parts. This means it struggles to understand the relationships among multiple parts (more than two parts).

\subsection{Variants of \MyModel}
We present three variants of \MyModel~in our main paper: \MyModel-CG, \MyModel-R, and \MyModel-T. As stated in the main paper, we modify the approach for modeling the score function for these variants. Specifically, we attempt to model the score function as $S_\theta = \nabla_c \log p_t(c_n^m \mid GNN(F_P^n, f_m))$, where $f_m$ represents the encoded feature for the $m^{th}$ part in the part set, and $c_n^m$ denotes the corresponding selection boolean value for this single part. The $GNN$ is implemented using an EdgeConv-based structure. \MyModel-CG, \MyModel-R, and \MyModel-T apply an MLP, ResNet \cite{he2016deep}, and Transformer \cite{vaswani2017attention}, respectively, to learn the new score function. In these variants, we separate the GNN from the score function, aiming to determine whether the \MyModel~framework can be effectively adapted to score functions modeled in this manner. Furthermore, we seek to explore whether applying better architectures can enhance the network's performance under this new modeling approach.

\section{Experimental details}
\paragraph{Training details} In our experiments, the optimizer applied for training is Adam \cite{kingma2014adam}. The learning rate is set as $10^{-3}$, and the batch size is set as 16. In the training procedure, we select the best checkpoints for each dataset.

\paragraph{Sampling details} As mentioned in our main paper, we use Predictor-Corrector sampler \cite{song2020score} for both selection vector sampling and pose matrix sampling. The parameters for both samplers are set as $T=1.0, \sigma=25.0, C=1$. The sampling step $N$ is set as $500$.

\section{Additional Experiments}

\begin{figure}[h]
    \centering
    \includegraphics[width=0.5\textwidth,trim=150 90 220 90,clip]{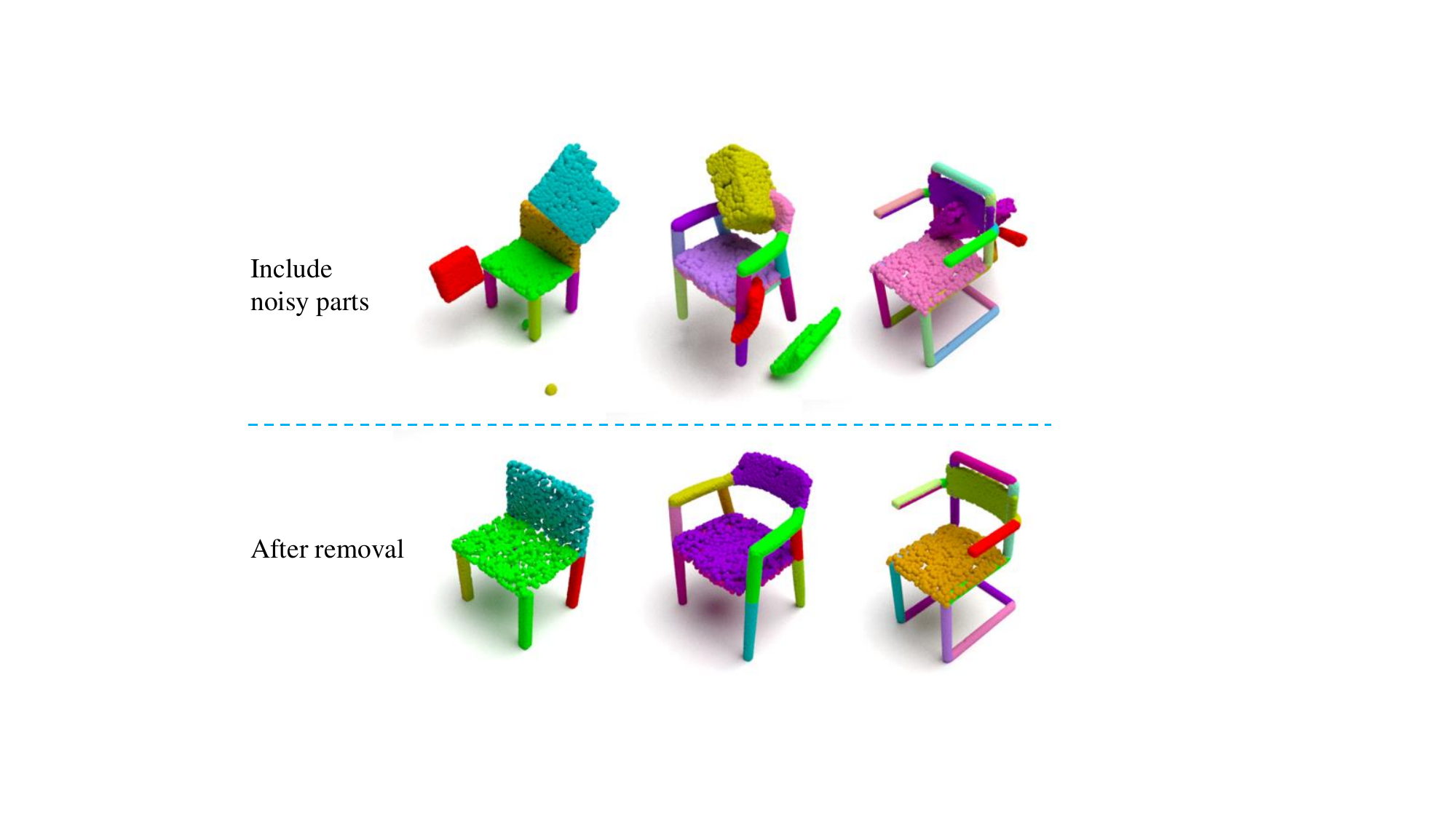}
    \caption{More qualitative results of noisy part removal task. Our framework can remove the unnecessary parts from the given set of parts. }
    
    \label{fig:part_complete}
\end{figure}

\subsection{Category Mixing Testing}
In this experiment, we mix all three categories (chair, table, and lamp) and test the performance of \MyModel~on the mixed-category dataset. The results are shown in Table~\ref{tab:mix_cats}. Surprisingly, we find that the performance on this mixed-category dataset is even better than that on single-category data. We infer that this improvement is due to two main reasons: 1. The mixing of categories results in a larger dataset, which may lead to better generalization; 2. Parts become more distinguishable when mixed together (e.g., chair parts versus lamp parts).
\begin{table}[h]
    
    \centering
    \small
    \begin{tabular}{ccc}
    \toprule
        Precision & Recall & F1 \\
        \midrule
         0.896 / 0.866& 0.804 / 0.792 & 0.84 / 0.827 \\
         \bottomrule
    \end{tabular}
    
    \caption{Results on all mixed categories data}
    
    \label{tab:mix_cats}
\end{table}

\subsection{Generalization to Unseen Categories}
To further verify the generalization ability of our algorithm for unseen category objects, we conducted an experiment, the results of which are shown in Table~\ref{tab:swap_test}. In this experiment, the model is trained on the chair dataset but tested on the table dataset. Although the performance is lower than that of the model trained and tested on the same table dataset, it is still capable of grouping parts from unseen categories. This demonstrates that our algorithm can generalize to a certain extent to object types it has not previously encountered.

\begin{table}[h]

    \centering
    \small
    \begin{tabular}{ccc}
         \toprule
        Precision & Recall & F1 \\
        \midrule
         0.766 / 0.716& 0.697 / 0.685 & 0.717 / 0.7 \\
         \bottomrule
    \end{tabular}
    
    \caption{Testing on Table with the model trained on Chair.}
    
    \label{tab:swap_test}
\end{table}

\subsection{More Qualitative Results}
In Fig.~\ref{fig:all_quali}, we present the full comparison for the Fig. 4 of the main paper. The full results indicate that the baseline methods struggle to accurately group the 3D parts. Besides, we also present additional qualitative comparisons in Fig.~\ref{fig:all_quali_2}. The figure shows that our framework is able to correctly group most part sets, while it is difficult for other baselines to obtain the correct groups. This result proves the effectiveness of our proposed method.

\subsection{Additional Results of Noisy Part Removal}
We demonstrate additional results of noisy part removal task in Fig.~\ref{fig:part_complete}. The figure shows that our framework can remove noisy parts from the given part sets in a zero-shot manner.

{
    \small
    \bibliographystyle{ieeenat_fullname}
    \bibliography{egbib}
}

\end{document}